%% file: SCC.tex
\documentclass[10pt,twocolumn,letterpaper]{article}

\usepackage{iccv}
\usepackage{times}
\usepackage{epsfig}
\usepackage{graphicx}
\usepackage{subfigure} 
\usepackage{amsmath}
\usepackage{amssymb}
\usepackage[utf8]{inputenc}
\usepackage[T1]{fontenc}

\usepackage{algorithm}
\usepackage{algorithmic}
\usepackage{amsfonts}
\usepackage{epstopdf}
\usepackage{amsthm}
\usepackage{xspace}
\usepackage{color}

\newcommand{\url}[1]{#1}

\newcommand{\xb}{\mathbf{x}}
\newcommand{\cb}{\mathbf{c}}

\newcommand{\hb}{\mathbf{h}}
\newcommand{\hbt}{\hat{\hb}}

\newcommand{\yt}{\hat{y}}

\newcommand{\Xb}{\mathbf{X}}
\newcommand{\Zb}{\mathbf{\hat X}}

\newcommand{\Tb}{\mathbf{T}}

\newcommand{\Bb}{\mathbf{B}}
\newcommand{\Mb}{\mathbf{M}}

\newcommand{\Ub}{\mathbf{U}}
\newcommand{\one}{\mathbf{1}}
\newcommand{\mub}{\boldsymbol{\mu}}

\newcommand{\set}[1]{\{#1\}}
\newcommand{\transpose}{\!^\top}

\newcommand{\Lc}{\mathcal{L}}
\newcommand{\Sc}{\mathcal{S}}

\newcommand{\balpha}{\boldsymbol{\alpha}}
\newcommand{\bbeta}{\boldsymbol{\beta}}

\newcommand{\w}{\mathbf{w}}

\newcommand{\knn}{{$k$NN}}
\newcommand{\BigO}[1]{\ensuremath{\operatorname{O}\bigl(#1\bigr)}}
\newcommand{\F}{\mathcal{F}}
\newcommand{\Fbb}{\boldsymbol{\F}}

\newcommand{\parakqw}[1]{

\vspace{1ex}\noindent\textbf{#1}}

\newtheorem{define}{Definition}

\DeclareMathOperator*{\argmin}{arg\,min}

\iccvfinalcopy 

\def\iccvPaperID{2697} 

\ificcvfinal\fi
\begin{document}

\title{Image Data Compression for Covariance and Histogram Descriptors}

\author{Matt J. Kusner, Nicholas I. Kolkin\\
Washington University in St. Louis \\
{\tt\small \{mkusner,n.kolkin\}@wustl.edu}
\and
Stephen Tyree\\
NVIDIA Research\\
{\tt\small styree@nvidia.com}
\and
Kilian Q. Weinberger\\
Washington University in St. Louis\\
{\tt\small kilian@wustl.edu}
}

\maketitle

\begin{abstract}
\input abstract

\end{abstract}

\section{Introduction}
\input{introduction.tex}

\section{Covariance and Histogram Descriptors}
\input{background.tex}\label{sec:background}

\section{Covariance compression}
\input{compression.tex}
\label{sec:compression}

\section{Histogram compression}
\input{hist_compression.tex}\label{sec:hist_compression}


\section{Results}
\input{results.tex}\label{sec:results}

\section{Related Work}
\input{related.tex}\label{sec:related}

\section{Conclusion}
\input{conclusion.tex}\label{sec:conclusion}

{\small
\bibliographystyle{ieee}
\bibliography{compressed}
}


\end{document}

%% file: abstract.tex
Covariance and histogram image descriptors provide an effective way to capture information about images.  
Both excel when used in combination with special purpose distance metrics.
For covariance descriptors these metrics measure the distance along the non-Euclidean Riemannian manifold of symmetric positive definite matrices.
For histogram descriptors the Earth Mover's distance measures the optimal transport between two histograms.
Although more precise, these distance metrics are very expensive to compute, making them impractical in many applications, even for data sets of only a few thousand examples.
In this paper we present two methods to compress the size of covariance and histogram datasets with only marginal increases in test error for $k$-nearest neighbor classification.
Specifically, we show that we can reduce data sets to $\mathbf{16}\boldsymbol{\%}$ and in some cases as little as $\mathbf{2}\boldsymbol{\%}$ of their original size, while approximately matching the test error of $k$NN classification on the full training set.
In fact, because the compressed set is learned in a supervised fashion, it sometimes even outperforms the full data set, while requiring only a fraction of the space and drastically reducing test-time computation.

%% file: introduction.tex
In the absence of sufficient data to learn image descriptors directly, two of the most influential classes of feature descriptors are (i) the histogram and (ii) the covariance descriptor. 
Histogram descriptors are ubiquitous in computer vision~\cite{belongie2002shape,dalal2005histograms,lazebnik2005sparse,lowe2004distinctive,mikolajczyk2005performance}.
These descriptors may be designed to capture the distribution of image gradients throughout an image, or they may result from a visual bag-of-words representation~\cite{fei2007recognizing,fei2005bayesian,leung2001representing}.
Covariance descriptors, and more generally, symmetric positive definite (SPD) matrices, are often used to describe structure tensors~\cite{goh2008clustering}, diffusion tensors~\cite{pennec2006riemannian} or region covariances~\cite{tuzel2006region}.
The latter are particularly well suited for the task of object detection from a variety of viewpoints and illuminations.


In this paper, we focus on the $k$-nearest neighbor ($k$NN) classifier with histogram or covariance image descriptors.
Computing a nearest neighbor or simply comparing a pair of histograms or SPD matrices is non-trivial.
For histogram descriptors, certain bins may be individually similar/dissimilar to other bins.
Therefore, the Euclidean distance is often a poor measure of distance as it cannot measure such bin-wise dissimilarity.
In the case of covariance descriptors, SPD matrices lie on a convex half-cone---a non-Euclidean Riemannian manifold embedded inside a Euclidean space.
Measuring distances between SPD matrices with the straight-forward Euclidean metric ignores the underlying manifold structure of the data and tends to systematically under-perform in classification tasks~\cite{vemulapalli2015riemannian}.

Histogram and covariance descriptors excel if their underlying structure is incorporated into the distance metric.  
Recently, there have been a number of proposed histogram distances~\cite{niblack1993qbic,pele2010quadratic,rubner2001empirical,stricker1995similarity}. Although these yield strong improvements in {\knn} classification accuracy (versus the Euclidean distance), these distances are often very costly to compute (\emph{e.g.} super-cubic in the histogram dimensionality). 
Similarly, for SPD matrices there are specialized geodesic distances~\cite{cherian2011efficient} and algorithms \cite{jayasumana2013kernel,tuzel2006region} developed that operate on the SPD covariance manifold. Because of the SPD constraint, these methods often require significantly more time to make predictions on test data. This is especially true for {\knn}, as the computation of the geodesic distance along the SPD manifold requires an eigen-decomposition \emph{for each individual pairwise distance}--- a computation that needs to be repeated for all training inputs to classify a single test input.

Cherian et al. \cite{cherian2011efficient} improved the running time (in practice) of test classification by approximating the Riemmanian distance with a symmetrized log-determinant divergence. For low dimensional data, Bregman Ball Trees \cite{cayton2008fast} can be adapted, however the performance deteriorates quickly as the dimensionality increases.


In this paper, we develop a novel technique to speed up $k$-nearest neighbor applications on covariance and histogram image features, one that can be used in concert with many other speedup methods. 
Our methods, called \emph{Stochastic Covariance Compression} (SCC) and \emph{Stochastic Histogram Compression} (SHC) learn a \emph{compressed training set} of size $m$, such that $m \!\ll\! n$, which approximately matches the performance of the {\knn} classifier on the original data. This new data set does not consist of original training samples; instead it contains new, artificially generated inputs which are explicitly designed for low {\knn} error on the training data. 
The original training set can be discarded after training and during test-time, as we only find the $k$-nearest neighbors among these artificial samples. This drastically reduces computation time and shrinks storage requirements. 
To facilitate learning compressed data sets we borrow the concept of stochastic neighborhoods, used in data visualization~\cite{hinton2002sne,vandermaaten2008visualizing} and metric learning~\cite{goldberger2004nca}, and leverage recent results from the machine learning community on data compression in Euclidean spaces~\cite{kusner2014stochastic}.

We make three novel contributions: 1. we derive SCC and SHC, two new methods for compression of covariance and histogram data; 2. we devise efficient methods for solving the SCC and SHC optimizations using the Cholesky decomposition and a normalized change of variable; 3. we carefully evaluate both methods on several real world data sets and compare against an extensive set of state-of-the-art baselines. 
Our experiments show that SCC can often compress a covariance data set to about $16\%$ of its original size, without increase in {\knn} test error.
In some cases, SHC and SCC can match the {\knn} test error with only $2\%$ of the training set size---leading to order-of-magnitude speedups during test time.
Finally, because we learn the compressed set explicitly to minimize {\knn} error, in a few cases it even \emph{outperforms} the full data set by achieving \emph{lower test error}.

%% file: background.tex
\label{sec:back}
We assume that we are given a set of $d$-dimensional feature vectors $\F = \set{\xb_1, \ldots, \xb_{|\F|}}\subset\mathbb{R}^d$ computed from a \emph{single} input image. From these features, we compute covariance or histogram image descriptors. 

\parakqw{Covariance descriptors} represent $\F$ through the covariance matrix of the features,
\begin{equation}
\Xb = \frac{1}{|\F|-1} \sum_{r=1}^{|\F|} (\xb_r - \mub)(\xb_r - \mub)\transpose,
\nonumber 
\end{equation}
where $\mub\!=\!\frac{1}{|\F|}\sum_{r=1}^{|\F|} \xb_r$. For vectorial data, `nearness' is often computed via the Euclidean distance or a learned Mahalanobis metric \cite{weinberger2009distance}. However, the Euclidean/Mahalanobis distance between two covariance matrices is a poor approximation to their true distance along the manifold of SPD matrices. A natural distance for covariance matrices is the Affine-Invariant Riemannian metric \cite{cherian2011efficient}, a geodesic distance on the SPD manifold. 

\begin{define}
\label{def:RM}
Let $\Sc_+^d$ be the positive definite cone of matrices of rank $d$. The \textbf{Affine-Invariant Riemannian metric} (AIRM) between any two matrices $\Xb, \Zb \in \Sc_+^d$ is $D_R(\Xb,\Zb) = \| \log(\Zb^{-1/2} \Xb \Zb^{-1/2}) \|_F$. 
\end{define}

While the AIRM accurately describes the dissimilarity between two covariances along the SPD manifold, it requires an \emph{eigenvalue decomposition for every input $\Zb$}. The metric becomes intractable to compute even for moderately-sized covariance matrices (\emph{e.g.,} computing $\Zb^{-1/2} \in \Sc_+^d$ requires roughly $\BigO{d^3}$ time). To alleviate this computational burden, a distance metric with similar theoretical properties has been proposed by \cite{cherian2011efficient}, called the Jensen-Bregman LogDet Divergence (JBLD),
\begin{align}\label{eq:jbld}
D_J(\Xb, \Zb) = \log \Big| \frac{\Xb + \Zb}{2} \Big| - \frac{1}{2} \log | \Xb \Zb |.
\end{align}
Cherian et al.~\cite{cherian2011efficient} demonstrate that for nearest neighbor classification, using JBLD as a distance has performance nearly identical to the AIRM but is much faster in practice and asymptotically requires $O(d^{2.37})$ computation~\cite{coppersmith1990matrix}.

\parakqw{Histogram descriptors}
are a popular alternative to covariance representations. Assume we again have a set of $d$-dimensional features for an image $\F = \{\mathbf{x}_1, \ldots, \mathbf{x}_{|\F|}\}$. Further, let the collection of all such features for all $n$ images in a training set be $\boldsymbol{\F} = \F_i \cup \ldots \cup \F_n$ (where $\F_i$ are the features for image $i$). To construct the visual bag-of-words representation we cluster all features in $\Fbb$ into $K$ centroids $\cb_1,\dots,\cb_K$ (e.g., via $k$-means), where these centroids are often referred to as a \emph{codebook} \cite{fei2005bayesian}. Using this codebook the visual bag-of-words representation $\hb_i$ of an image $i$ is a $K$-dimensional vector, where element $h_{ij}$ is a count of how many features in the bag $\F_i$ have $\cb_j$ as the nearest centroid.

Arguably one of the most successful histogram distances is the Earth Mover's Distance (EMD) \cite{rubner2000earth}, which has been used to achieve impressive results for image classification and retrieval \cite{levina2001earth,ling2007efficient,rubner2000earth,pele2009fast}. 
EMD constructs a distance between two histograms by `lifting' a bin-to-bin distance, called the \emph{ground distance} $\Mb$, where $M_{ij}\!\geq\! 0$, to a full histogram distance. Specifically, for two histogram vectors $\hb$ and $\hb'$ the EMD distance is the solution to the following linear program:
\begin{equation}
\min_{\Tb \geq 0}\  \textrm{tr}(\Tb \Mb) \; \textrm{ s.t. } \Tb\mathbf{1}=\hb \textrm{ and } \Tb^\top\mathbf{1}=\hb'\label{opt:emd},
\end{equation}
where $\Tb$ is the transportation matrix and $\one$ is a vector of ones. Each element $T_{ij}$ describes the amount of mass moved from $h_i$ to $h_j'$, for the vectors to match exactly. 
One example ground distance for the visual bag-of-words representation is the Euclidean distance between the centroid vectors $M_{ij} \triangleq \| \cb_i - \cb_j \|_2$. When the ground distance is a metric, it can be shown that the EMD is also a metric \cite{rubner2000earth}.

In practice, one limitation of the EMD distance is its high computational complexity. Cuturi et al. \cite{cuturi2013sinkhorn} therefore introduce the Sinkhorn Distance, which involves a regularized version of the EMD optimization problem:
\begin{align}
\min_{\Tb \geq 0}  \textrm{tr}(\Tb\Mb) - \frac{1}{\lambda} h(\Tb), \textrm{ s.t. } \Tb\mathbf{1}=\hb \textrm{ and } \Tb^\top\mathbf{1}=\hb' \label{eq:primal1}
\end{align}
where 
$h(\Tb)\!=\!-\textrm{tr}(\Tb \log(\Tb))$
is the entropy of the transport $\Tb$. The Sinkhorn distance between $\hb$ and $\hb'$ is $D_S(\hb,\hb') =  \textrm{tr}(\Tb^\lambda\Mb)$, where $\Tb^\lambda$ is the solution to \eqref{eq:primal1}. The solution is an arbitrarily close upper bound to the exact EMD solution (by increasing $\lambda$) and the optimization problem is shown to be at least an order of magnitude faster to compute than the EMD linear program \eqref{opt:emd}. Specifically, Cuturi et al. introduce a simple iterative algorithm to solve eq.(\ref{eq:primal1}) in time $\BigO{d^2 i}$, where $d$ is the size of the histograms and $i$ is the number of iterations of the algorithm. This is compared to $\BigO{d^3 \log d}$ complexity of the EMD optimization problem \cite{pele2009fast}. In practice, each algorithm iteration is a matrix scaling computation that can be performed between multiple histograms simultaneously. This means that the algorithm is parallel and can be efficiently computed on modern hardware architectures (\emph{i.e.}, multi-core CPUs and GPUs) \cite{cuturi2013sinkhorn}. 

%% file: compression.tex
In this section we detail our covariance compression technique: \emph{Stochastic Covariance Compression} (SCC). SCC uses a stochastic neighborhood to compress the training set from $n$ input covariances to $m$ `compressed' covariances. After learning, the original training set can be discarded and all future classifications are made just using the compressed inputs. Since $m \ll n$, the complexity of test-time classification is drastically reduced, from $\BigO{nd^{2.37}}$ to $\BigO{md^{2.37}}$, where $\BigO{d^{2.37}}$ is the asymptotic complexity of computing a single JBLD distance in eq.~(\ref{eq:jbld}).

Assume we are given a training set of $n$ covariance matrices $\{\Xb_1,\dots ,\Xb_n\}\!\subset\! {\mathbb{R}}^{d\times d}$ with corresponding labels $y_1, \ldots, y_n$. 
Our goal is to learn a compressed set of $m$ covariance matrices $\{\Zb_1,\dots,\Zb_m\} \! \subset\! \mathbb{R}^{d\times d}$ with labels $\hat{y}_1, \ldots, \hat{y}_m$. To initialize $\Zb_j$, we randomly sample $m$ covariance matrices from our training data set and copy their associated labels for each $\hat{y}$.
We optimize these synthetic inputs $\Zb_j$ to minimize the {\knn} classification error.
The {\knn} classification error is non-continuous and non-differentiable with respect to $\Zb_j$, but we can introduce a stochastic neighborhood, as proposed by Hinton and Roweis~\cite{hinton2002sne}, to ``soften'' the neighborhood assignment and allow optimization on {\knn} error.
Specifically, we place a radial basis function around each input $\Xb_i$ and proceed as if the nearest prototypes $\Zb_j$ are assigned randomly.
For a given $\Xb_i$,  the probability that $\Zb_j$ is picked as the nearest neighbor is denoted
\begin{align} \label{eq:cov_stoch_neigh}
p_{ij} = \frac{ e^{-\gamma^2 D_J(\Xb_i,\Zb_j) } }{ \sum_{k=1}^m e^{-\gamma^2 D_J(\Xb_i,\Zb_k) } } = \frac{1}{\Omega_i}  e^{-\gamma^2 D_J(\Xb_i,\Zb_j)},
\end{align}
where $D_J(\Xb_i,\Zb_j)$ is the JBLD divergence in eq.~(\ref{eq:jbld}) and $\Omega_i$ denotes the normalization term.
The constant $\gamma\!>\!0$ is a hyper-parameter defining the ``sharpness'' of the neighborhood distribution. (We set $\gamma$  by cross-validation) 



\parakqw{Objective.}
Inspired by Neighborhood Components Analysis~\cite{goldberger2004nca}, we can compute the probability $p_i$ that an input $\Xb_i$ will be classified correctly by this stochastic nearest neighbor classifier under the compressed set $\{\Zb_1, \ldots, \Zb_m\}$, 
\begin{align} \label{eq_goldberger_pi}
	p_i = \sum_{j: y_j = y_i} p_{ij}.
\end{align}
Ideally, $p_i\!=\!1$ for all $\Xb_i$, implying the compressed set yields perfect predictions on the training set. The KL-divergence between this ideal ``1-distribution'' and $p_i$ is simply $KL(1 || p_i) = -\log(p_i)$. Our objective is to minimize the sum of these KL-divergences with respect to our compressed set of covariance matrices $\{\Zb_1, \ldots, \Zb_m\}$,
\begin{align}\label{eq:obj}
\min_{\{\Zb_1, \ldots, \Zb_m\}} - \sum_{i=1}^n \log(p_i).
\end{align}

\parakqw{Gradient.}
To ensure that the learned matrices $\Zb_j$ are SPD, we decompose each matrix $\Zb_j$ by its unique Cholesky decomposition: $\Zb_j\!=\!\Bb_j^\top\Bb_j$, where $\Bb_j$ is an upper triangular matrix. To ensure that $\Zb_j$ remains SPD we perform  gradient descent w.r.t. $\Bb_j$. The gradient of ${\cal L}$ w.r.t. $\Bb_j$ is
\begin{align}
\frac{\partial \Lc}{\partial \Bb_j} = \sum_{i=1}^n \frac{p_{ij}}{p_i} (\delta_{y_iy_j} - p_i)  \gamma^2 \frac{\partial D_J(\Xb_i, \Bb_j^\top\Bb_j)}{\partial \Bb_j} \label{eq:gradB}
\end{align}
where $\delta_{y_iy_j} \!=\! 1$ if $y_i \!=\! y_j$ and is $0$ otherwise and $D_J(\Xb_i, \Bb_j^\top\Bb_j)$ is the JBLD divergence between $\Xb_i$ and $\Bb_j^\top\Bb_j = \Zb_j$. The gradient of the JBLD w.r.t. $\Bb_j$ is:
\begin{align}
\frac{\partial D_J(\Xb_i, \Bb_j^\top\Bb_j) }{ \partial \Bb_j} \!=\! \Bb_j( \Xb_i \!+\!  \Bb_j^\top \Bb_j)^{-1} \!-\!  (\Bb_j^{\top})^{-1}. \label{eq:grB} 
\end{align}
We substitute~\eqref{eq:grB} into \eqref{eq:gradB} to obtain the final gradient. 
For a single compressed input $\Zb_j =  \Bb_j^\top\Bb_j$, each step of gradient descent requires $\BigO{ d^3 }$ to compute $\frac{\partial D_J(\Xb_i, \Bb_j^\top\Bb_j)}{\partial \Bb_j}$ and $\BigO{d^{2.37}}$ to compute $D_J(\Xb_i, \Zb_j)$. 
It requires $\BigO{d^3 m}$ to compute $p_{ij}$ and an additional $\BigO{m}$ for $p_i$.
Thus the overall complexity of $\frac{\partial {\cal L}}{\partial \Zb_j}$ is $\BigO{d^3 m^2 n}$. 
We minimize our objective in eq.~(\ref{eq:obj}) via conjugate gradient descent.\footnote{\url{http://tinyurl.com/minimize-m}}
A Matlab implementation of SCC is available at: \url{http://anonymized}.

%% file: hist_compression.tex
Analogous to covariance compression, we can also compress histogram descriptors, which we refer to as \emph{Stochastic Histogram Compression} (SHC).
Our aim is to learn a compressed set of $m \ll n$ histograms $\{\hbt_1, \ldots, \hbt_m\} \!\subset\! {\boldsymbol{\Sigma}_d}$ with labels $\yt_1, \ldots, \yt_m$ from a training set histograms $\{\hb_1, \ldots,\hb_n\} \!\subset\! {\Sigma_d}$ with labels $y_1, \ldots, y_n$, where $\boldsymbol{\Sigma}_d$ is the $(d\!-\!1)$-dimensional simplex.

\parakqw{Objective.}
As before we place a stochastic neighborhood distribution over compressed histograms and define the probability that $\hbt_j$ is the nearest neighbor of $\hb_i$ via 
\begin{align} \label{eq:pij_hist}
p_{ij} = \frac{1}{\Omega_i}  e^{-\gamma^2 D_S(\hb_i,\hbt_j) } 
\end{align}
where $D_S$ is the Sinkhorn distance and $\Omega_i$ normalizes $p_{ij}$ so that it is a valid probability. As in SCC, we define the probability that a training histogram $\hb_i$ is predicted correctly as $p_i$ via eq.~(\ref{eq_goldberger_pi}) by summing over the compressed inputs $\hbt_k$ the share the same label.
We then minimize the KL-divergence between the perfect distribution and $p_i$ as in eq.~(\ref{eq:obj}) to learn our set of compressed histograms $\hbt_j$.

\parakqw{Gradient.}
As in the covariance setting, the gradient of the objective in eq.~(\ref{eq:obj}) w.r.t. a compressed histogram $\hbt_j$ is
\begin{align}
\frac{\partial \Lc}{\partial \hbt_j} = \sum_{i=1}^n \frac{p_{ij}}{p_i} (\delta_{y_iy_j} - p_i)  \gamma^2 \frac{\partial D_S(\hb_i, \hbt_j)}{\partial \hbt_j}. \label{eq:shc_grad}
\end{align}
The gradient of the Sinkhorn distance $D_S$ w.r.t. $\hbt_j$ introduces two challenges: 1. the distance $D_S$ itself is a nested optimization problem; and 2. the learned vector $\hbt_j$ must remain a well-defined histogram throughout the optimization, \emph{i.e.} it must be non-negative and sum to $1$, s.t. $\hbt_j \in \boldsymbol{\Sigma}_d$. 

We first address the gradient of the nested optimization problem w.r.t. $\hbt_j$, \emph{i.e.} $\frac{\partial D_S(\hb_i, \hbt_j)}{\partial \hbt_j}$. 
In the \emph{primal} formulation, as stated in eq.~(\ref{eq:primal1}), the histogram $\hbt_j$ occurs within the constraints, which complicates the gradient computation. Instead, we form the \emph{dual}~\cite{cuturi2013sinkhorn}, 
\begin{align}
\max_{\balpha,\bbeta \in \mathbb{R}^d} \balpha^\top \hb_i + \bbeta^\top \hbt_j - \sum_{k,l = 1}^d \frac{e^{-\lambda(M_{ij} - \alpha_i - \beta_j)}}{\lambda},\label{eq:dual}
\end{align}
where $\balpha,\bbeta$ are the corresponding dual variables.
Due to strong duality, the primal and dual formulations are identical at the optimum, however the dual formulation (\ref{eq:dual}) is \emph{unconstrained}. 
The gradient of the dual objective (\ref{eq:dual}) is \emph{linear} w.r.t. $\hbt_j$. If we consider $\bbeta$ fixed, it follows that at the optimum $\frac{\partial D_S(\hb_i, \hbt_j)}{\partial \hbt_j}\approx \bbeta^*$, where  $\bbeta^*$ the optimal value of $\bbeta$. This optimal dual variable is easily computed with the iterative Sinkhorn algorithm~\cite{cuturi2014fast} mentioned in Section~\ref{sec:back}. This approximation ignores that $\bbeta^*$ itself is a function of $\hbt_j$, which is a reasonable approximation for small step-sizes. 

To address the second problem and simultaneously perform (approximated) gradient descent while ensuring that $\hbt_j$ always lies in the simplex (i.e., is normalized),
we propose a change of variable in which we redefine each compressed histogram $\hbt_j$ as a positive, normalized quantity: $\hbt_j = e^{ \w_j } / \sum_{k=1}^d e^{ \w_{jk} }$ for $\w_j \in \mathbb{R}^d$.
Then the gradient of the Sinkhorn distance can be taken with respect to $\w_j$, 
\begin{align}
\frac{\partial D_S(\hb_i, \hbt_j)}{\partial \w_j} \approx \bbeta^* \circ \Big( \frac{ s e^{\w_j} - e^{2 \w_j}}{s^2} \Big) \nonumber
\end{align}
where $s$ is the normalizing term $s = \sum_{k=1}^d e^{ \w_{jk} }$ and $\circ$ is the Hadamard (element-wise) product. The complexity of computing the full SHC gradient in eq.~(\ref{eq:shc_grad}) is $\BigO{nmd^3 \hat{i}}$: each Sinkhorn gradient above requires time $\BigO{d^3 \hat{i}}$ (where $\hat{i}$ is the number of Sinkhorn iterations), and $p_{ij}$ requires time $\BigO{\hat{i} d^2 m}$. 
We use gradient descent with an updating learning rate to learn $\hbt_j$, selecting the compressed set that yields the best training error across all iterations. 

\begin{figure*}[t]
\begin{center}
\centerline{\includegraphics[width=\textwidth]{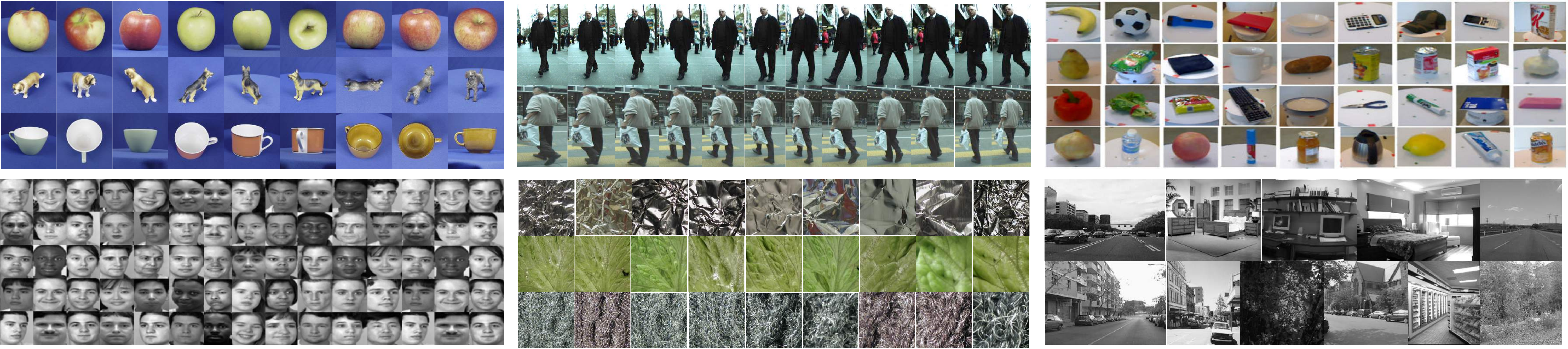}}
\caption{Montages of the $6$ datasets used in our evaluations from top down, left to right: (a) {\sc{eth80}} objects at different orientations; (b) {\sc{ethz}} person recognition; (c) {\sc{rgbd}} objects from point clouds; (d) {\sc{feret}} face detection; (e) {\sc{kth-tips2b}} material categorization (f) {\sc{scene15}} scene classification.}
\label{fig:data}
\end{center}
\vspace{-5ex}	
\end{figure*}

%% file: results.tex

We evaluate both algorithms on a series of real-world data sets and compare with state-of-the-art algorithms for {\knn} compression. 

\subsection{Covariance compression}

\parakqw{Datasets.} We evaluate our covariance compression method on six benchmark data sets. The {\sc{eth80}} dataset has images of $8$ object categories, each pictured with a solid blue background. For each category there are $10$ exemplar objects and for each exemplar the camera is placed in $41$ different positions. 
We use the $19 \times 19$ covariance descriptors of \cite{cherian2014riemannian}, who segment the images and use per-pixel color and texture features to construct covariances. 
The {\sc{ethz}} dataset is a low-resolution set of images from surveillance cameras of sizes from $78 \times 30$ to $400 \times 200$. The original task is to identify the person in a given image, from $146$ different individuals. The original dataset has multiple classes with fewer than $10$ individuals. Therefore, to better demonstrate a wide range of compression ratios, we filter the dataset to include only the most popular $50$ classes resulting in each individual having between $59$ and $356$ images ($5193$ images total). We use the pixel-wise features of \cite{cherian2014riemannian} to construct $18 \times 18$ covariance matrices. 
The {{\sc{feret}}} face recognition dataset has $3737$ gray-scale images of the faces of $399$ individuals, oriented at various angles. As the majority of individuals have fewer than $5$ images in the training set, we also limit the dataset to the $50$ most popular individuals and use a larger set of compression ratios (described further in the error analysis subsection). We use the $40 \times 40$ Gabor-filter covariances of \cite{cherian2011efficient}. 
Our version of the {\sc{rgbd}} Object dataset \cite{lai2011large} contains $15,000$ point cloud frames of objects from three different views. The task is to classify an object as one of $51$ object categories. We use the $18 \times 18$ covariance features of \cite{cherian2014riemannian} which consist of intensity and depth-map gradients, as well as surface normal information. 
The \emph{{\sc{scene15}}} data consists of $4485$ black and white images of $15$ different indoor and outdoor scenes. We split the dataset into training and test sets as per \cite{lazebnik2006beyond}. To create covariance features we compute a dense set of SIFT descriptors centered at each pixel in the image\footnote{We use the open source library VLFeat http://www.vlfeat.org/}. 
Our SIFT features have $4$ bins each in the horizontal and vertical directions and $8$ orientation bins, producing a $128 \times 128$ covariance descriptor. Via the work of \cite{harandi2014manifold} we can learn a rank-$r$ projection matrix $\Ub \in \mathbb{R}^{d \times r}$, where $r \ll d$ to reduce the size of these covariance matrices to $r \times r$ via the transformation $\Xb_i \rightarrow \Ub^\top \Xb_i \Ub$. We use Bayesian optimization \cite{gardner2014bayesian}\footnote{https://bitbucket.org/mlcircus/bayesopt.m}, to select values for the covariance size $r$, as well as two hyperparameters in \cite{harandi2014manifold}: $\nu_v$ and $\nu_w$, by minimizing the $1$-NN error on a small validation set. 
The {\sc{kth-tips2b}} dataset is a material classification dataset of $11$ materials with $4752$ total images. Each material has $4$ different samples each from $3$ fixed poses, $9$ scales, and $4$ lighting conditions. We follow the procedure of \cite{harandi2014bregman} to extract $23 \times 23$ covariance descriptors using color information and Gabor filter responses.



\parakqw{Experimental Setup.}
We compare all methods against test error of $1$-nearest neighbor classification that uses the entire training set. For results that depend on random initialization or sampling we report the average and standard deviation across $5$ random runs (save {\sc{kth-tips2b}}, for which we use $4$ splits by holding out each of the four provided samples, one at a time). For datasets {\sc{rgbd}}, {\sc{ethz}}, and {\sc{eth80}} we report results averaged over $5$ different train/test splits.

\parakqw{Baselines.}
We compare our method, \emph{Stochastic Covariance Compression} (SCC), against a number of methods aimed at reducing the size of the training set, which we adapt for the covariance feature setting: 1. \emph{$k$NN} using the full training set, 2. $k$NN using a class-based \emph{subsampled} training set, which we use as initialization for SCC, 3. \emph{Condensed Nearest Neighbor} (CNN) \cite{hart1968condensed}, 4. \emph{Reduced Nearest Neighbor} (RNN) \cite{gates1972reduced}, 5. \emph{Random Mutation Hill Climbing} (RMHC) \cite{skalak1994prototype}, and 6. \emph{Fast CNN} (FCNN) \cite{icml2005_Angiulli05}.
Both CNN and FCNN select subsets of the training set that have the same leave-one-out training error as the full training set, and are very well-known in the fast $k$NN literature. RNN works by post-processing the output of CNN to improve the training error and RMHC is a random subset selection method. (We give further details on these algorithms in Section~\ref{sec:related}). For FCNN we must make a modification to accommodate covariance matrix features. Specifically, FCNN requires computing the centroid of each class at regular intervals during the selection. A centroid of class $y$ is given by solving the following optimization,
\begin{align}
\Xb_y = \argmin_{\Xb}  \sum_{i : y_i = y} D_J(\Xb,\Xb_i) \nonumber
\end{align}
where $D_J(\Xb, \Xb_i)$ is the JBLD divergence. Cherian et al.~\cite{cherian2011efficient} give an efficient iterative procedure for solving the above optimization, which we use in our covariance FCNN implementation.

\begin{figure*}[t]
\vspace{-2ex}
\begin{center}
\centerline{\includegraphics[width=0.85\textwidth]{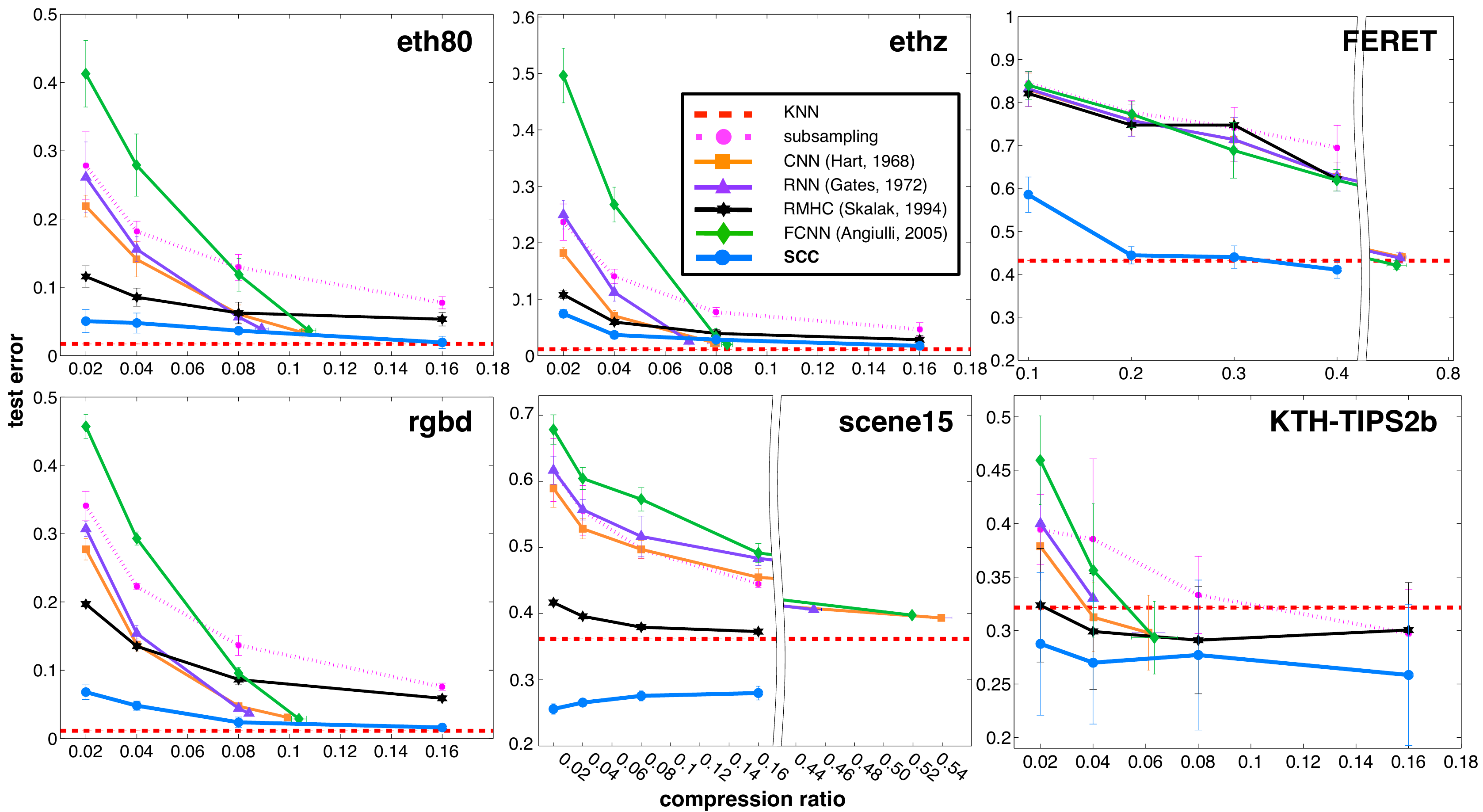}}
\caption{\knn{} test error rates after training set compression for covariance descriptors. See text for details.  }
\label{figure.compare}
\end{center}
\vspace{-5ex}
\end{figure*}

\parakqw{Classification error.}
Figure \ref{figure.compare} compares the test error of SCC to baselines for sizes of the compressed set equal to $2\%, 4\%, 8\%,$ and $16\%$ of the training set. For dataset {\sc{feret}} which has a large number of classes we use larger compression ratios: $10\%, 20\%, 30\%,$ and $40\%$. Although CNN, RNN, and FCNN only output a single reduced training set (the final point on each curve) we plot the intermediate test errors of each method at the above compression ratios as well. On each dataset SCC is able to reduce the test error to nearly that of $k$NN applied to the full dataset using less than or equal to $20\%$ of the training data. Only on {\sc{ethz}} and {\sc{rgbd}} could SCC not match the full $k$NN error up to significance, however the error rates are only marginally higher. For small compression ratios SCC is superior to all of the baselines, as well as the subsampling initialization. On datasets {\sc{eth80}} and {\sc{ethz}} the final outputs of CNN, RNN and FCNN are roughly equivalent to the SCC curve. However, one notable downside is that these algorithms have no control on the size of these final sets, which for {\sc{feret}} are as large as $75\%$ (CNN/RNN) and $74\%$ (FCNN). In contrast SCC allows one to regulate the compressed set size precisely. RMHC is also able to regulate the compressed set size. However, because it is based on random sampling, its performance can be very poor, as on {\sc{feret}}. 
Surprisingly, for {\sc{scene15}} and {{\sc{kth}}, learning a compressed covariance dataset with SCC reduces the test error \emph{below the $k$NN error of the full training set}. We suspect that this occurs because (a) the training set may have some amount of label noise, which is partially alleviated by subsampling, and (b) SCC essentially learns a new, supervised covariance representation, versus a label-agnostic set of covariance descriptors. For these datasets, there is no reason not to shrink the dataset to only $2\%$ of its original size and discard the original data---yielding $36\times$ and $48\times$ speedups during test-time.  


\parakqw{Test-time speedup.}
Table \ref{table.speedup} shows the speedup of SCC over $k$NN classification using the full training set for various compression ratios (the datasets marked by a (C) are learned with SCC). In general the speedups are roughly $1/\delta$, where $\delta$ denotes the compression ratio. Results that match or exceed the accuracy (up to significance) are in \textcolor{blue}{blue}. At $16\%$ compression $3$ of the $5$ datasets run at this compression ratio match the test error of full $k$NN classification. In effect we have removed neighbor redundancies in the dataset, and gained a factor of roughly $6\times$ speedup. Much larger speedups can be obtained at $4\%$ or $2\%$ compression ratio---although at a small increase in classification error. For the data set with many classes ({\sc{feret}}) ``loss-free'' compression can still yield a speedup of $5\times$ at $\delta=0.2$. 


\input{table_speedup.tex}

\parakqw{Training time.}
Table \ref{table.training} describes the average training times for SCC (again (C) denotes SCC results). For maximum compression to $2\%$ the training time is on the order of minutes. As the size of the compressed set gets larger the time increases but only by small amounts, indeed the longest training time is $2$ hours for {\sc{rgbd}} with $16\%$ compression. Furthermore, the entire compression can be done completely off-line prior to testing. 
The contributions of the training points to the gradient are independent and have a high computation to memory load ratio.  The SCC training could therefore potentially be sped up significantly through parallelization on clusters or GPUs. 

\begin{table}[t]
\caption{SCC and SHC training times.}
\label{table.training}
\begin{center}
\resizebox{8cm}{!} {
\begin{tabular}{ccccc}
\hline
\multicolumn{5}{c}{{\sc{\textbf {Training Times}}}}\\
\hline
\sc{Dataset} & \multicolumn{4}{c}{\sc{Compression Ratio}}  \\ 
(few classes)  & 2\% & 4\% & 8\% & 16\% \\
\hline
\sc{eth80} (C) & $1$m $9$s & $1$m $46$s & $2$m $39$s & $4$m $36$s \\
\sc{ethz} (C) & $2$m $4$s & $3$m $15$s & $5$m $53$s & $10$m $10$s \\
\sc{rgbd} (C) & $18$m $26$s & $33$m $30$s & $1$h $4$m & $2$h $11$m \\
\sc{scene15} (C) & $1$m $18$s & $1$m $52$s & $2$m $42$s & $4$m $20$s \\
\sc{kth} (C) & $3$m $31$s & $5$m $48$s & $9$m $2$s & $16$m $18$s \\
\hline
\sc{coil20} (H) & $15$s & $3$m $21$s & $4$m $43$s & $9$m $17$s \\
\sc{kylberg} (H) & $1$m $41$s & $2$m $46$s & $5$m $24$s & $11$m $17$s \\

\hline
\hline

\sc{Dataset} & \multicolumn{4}{c}{\sc{Compression Ratio}} \\
(many classes) & 10\% & 20\% & 30\% & 40\% \\
\hline
\sc{feret} (C) & $1$m $39$s & $2$m $19$s & $2$m $44$s & $3$m $6$s \\
\hline
\sc{mpeg7} (H) & $58$s & $19$m $26$s & $29$m $33$s & $38$m $52$s \\
\hline
\end{tabular}
}
\end{center}
\end{table}

\subsection{Histogram compression}

\begin{figure*}[t]
\begin{center}
\centerline{\includegraphics[width=\textwidth]{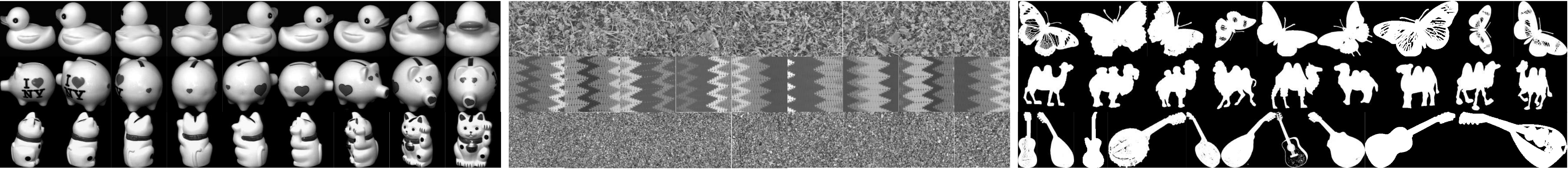}}
\caption{Montages of datasets used in SHC evaluation from left to right: (a) {\sc{coil20}} 3D object recognition; (b) {\sc kylberg} texture classification; (c) {\sc MPEG7} shape detection.}
\label{fig:data}
\end{center}
\vspace{-3ex}	
\end{figure*}

We evaluate our technique for compressing histogram datasets, Stochastic Histogram Compression (SHC) against current baseline methods for constructing a reduced training set. As a benchmark, we compare the $k$-nearest neighbor accuracies for compressed sets of different sizes, and report the test-time speedups achieved by our method. We start by describing the datasets we use for comparison.

\parakqw{Datasets.}
The {{\sc{coil20}}} dataset consists of $20$ grayscale image objects with background masked out in black. Each object was rotated $360$ degrees and an image was taken every $5$ degrees, yielding $72$ images per class. To construct histogram features we follow the procedure of \cite{belongie2002shape} to extract shape context log-polar histograms using $100$ randomly sampled edge points, yielding histograms of dimensionality $d\!=\!60$. As a ground distance $\Mb$ we use the $\ell_1$ distance between bins of the log-polar histogram.
The {{\sc{mpeg7}}} dataset has $70$ different shape classes, each with $20$ images. Each image has a black background with a sold white shape such as \emph{bat}, \emph{cellular phone}, \emph{fountain}, and \emph{octopus}, among others. We follow the procedure for the {\sc{coil20}} dataset to extract shape context histograms, also used in \cite{belongie2002shape} for the {\sc{mpeg7}} dataset. The ground distance is also the $\ell_1$ between bins.
The {\sc{kylberg}} texture dataset is a $28$-class dataset of different surfaces. We used the dataset without rotations that contains $160$ images for each class. We follow the feature-extraction technique of \cite{harandi2014bregman}, which uses first and second order image-gradient features at every 4 pixels, after resizing. We then construct a visual bag-of-words representation by first clustering all features into $50$ codewords. We represent each image as a $50$-dimensional count vector: the $i^{th}$ entry corresponds to the number of times a gradient feature was closest (in the Euclidean sense) to the $i^{th}$ codeword. As a ground distance between bins we use the Euclidean distance between each pair of codewords.


\parakqw{Experimental setup.}
As for covariance features, our benchmark for comparison of all methods is the test error of $1$-nearest neighbor classification with the full training set. Similarly, for each dataset we report results over $5$ different train/test splits. For our algorithm, SHC, we use Bayesian optimization \cite{gardner2014bayesian} to tune the $\gamma^2$ parameter in the definition of $p_{ij}$, eq.~(\ref{eq:pij_hist}), as well as the initial gradient descent learning rate, to minimize the training error. Additionally, we initialize SHC with the results of RMHC, which in the covariance setting appears to largely outperform the subsampling approach. We use the exact same baselines for covariance features, except now we use the Sinkhorn distance as our dissimilarity measure. The only subtlety is that FCNN needs to be able to compute the centroid of a set of histograms, with respect to the Sinkhorn distance. The centroid of a set of histogram measures with respect to the EMD is called the \emph{Wasserstein Barycenter} \cite{cuturi2014fast}. It is shown how to compute this barycenter for the Sinkhorn distance in \cite{cuturi2013sinkhorn}, and we use their accelerated gradient approach to solve for each Sinkhorn centroid.

\begin{figure*}[t]
\vspace{-2ex}
\begin{center}
\centerline{\includegraphics[width=0.85\textwidth]{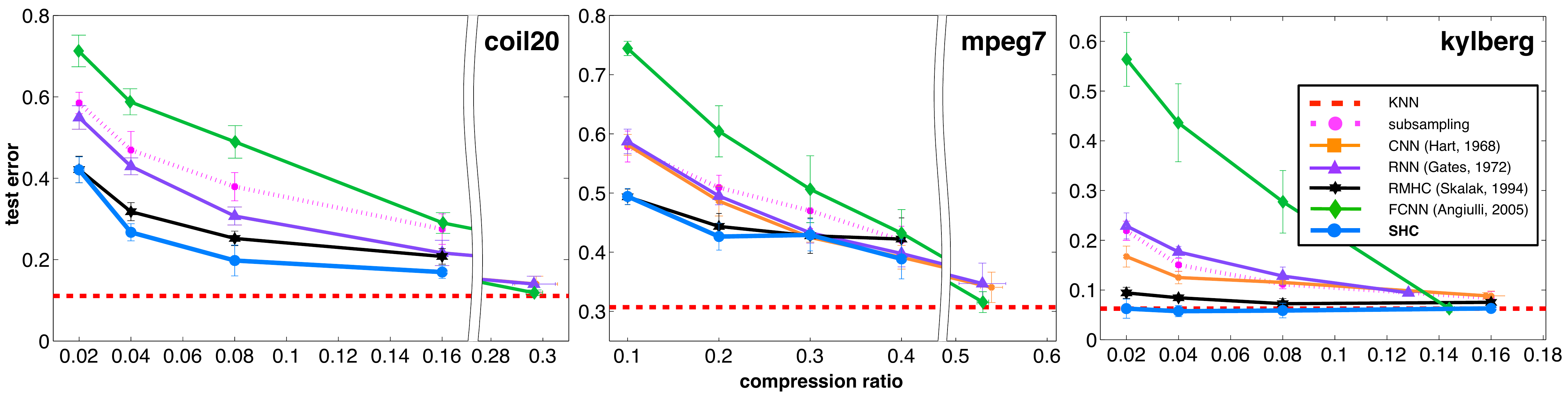}}
\caption{\knn{} test error rates after training set compression for histogram descriptors. See text for details.  }
\label{figure.compare_hist}
\end{center}
\vspace{-5ex}
\end{figure*}


\parakqw{Classification error.}
Figure~\ref{figure.compare_hist} shows the average test error and standard deviation for compression ratios of $2\%, 4\%, 8\%,$ and $16\%$ for {\sc{coil20}} and {\sc{kylberg}} and $10\%, 20\%, 30\%, 40\%$ for the many-class dataset {\sc{mpeg7}}. for each of the above datasets. As in the covariance setting, SHC outperforms or matches the error vs. compression trade-offs of all of the baseline methods throughout all evaluated settings. 
For a specific speedup over full $k$NN classification (\emph{i.e.} compression ratio), SHC is able to achieve the lowest test error (possibly matched by other methods) throughout. On {\sc {kylberg}}, SHC can reduce the training set to $1/50$ of its size without an increase in test error. 
%
%
%
On {\sc{coil20}} and {\sc{mpeg7}} the final compressed sets of CNN, RNN and FCNN have very high compression ratio (around $0.3$ and $0.5$), which lead to only very modest speedups. We did not evaluate SHC in these arguably least interesting settings, but nevertheless show the error rates of the baselines for completeness. 
%
%
%

\parakqw{Test-time speedup.}
The {\knn} test time speedups of SHC over the full training set are shown in Table \ref{table.speedup} (the (H) datasets). Similar to SCC, the speedups reach up to $46.9\times$ at maximum compression and still reach an order of magnitude in the worst case ($10\times$) on the {\sc mpeg7} data set with many classes. For {\sc{kylberg}} the SHC error is lower than the full data set---even at a $2\%$ compression ratio.

\parakqw{Training time.}
The training times of SHC are shown in Table \ref{table.training}. SHC is very fast ($<\!2$ minutes for $2\%$ compression on {\sc{kylberg}}), especially considering that we are solving a nested optimization problem over the compressed histograms and the Sinkhorn distance. 
We believe that the speed of our implementation can be further improved with the use of GPUs for the Sinkhorn computation and with approximate second-order hill-climbing methods. 

%% file: table_speedup.tex
\begin{table}
\caption{Speed-up of {\knn} testing through SCC and SHC compression. The SCC datasets are denoted with a (C) and the SHC datasets with an (H). Results where SCC/SHC matches or exceeds the accuracy of full \knn{} (up to statistical significance) are in \textcolor{blue}{blue}.}
\label{table.speedup}
\begin{small}
\begin{sc}
\resizebox{\columnwidth}{!} {
\begin{tabular}{c|c|c|c|c}
\hline
\multicolumn{5}{c}{\textbf{Speed-up}}\\
\hline
 Dataset & \multicolumn{4}{c}{Compression Ratio} \\ 
(few classes) & 2\% & 4\% & 8\% & 16\% \\
\hline
eth80 (C) & $47.6 \pm 0.1$ & $24.5 \pm 0.1$ & $12.3 \pm 0.2$ & $\textcolor{blue}{6.2 \pm 0.02}$ \\
ethz (C) & $48.1 \pm 1.0$ &	$24.5 \pm 0.1$ & $12.4 \pm 0.05$ &	$6.2 \pm 0.01$ \\
rgbd (C) & $34.0 \pm 0.8$ & $20.3 \pm 0.2$ & $10.9 \pm 0.2$ & $6.8 \pm 2.5$ \\
scene15 (C) & $\textcolor{blue}{36.0 \pm 0.6}$ & $\textcolor{blue}{20.6 \pm 0.3}$ & $\textcolor{blue}{11.3 \pm 0.1}$ & $\textcolor{blue}{5.9 \pm 0.08}$ \\
kth (C) & $\textcolor{blue}{43.0 \pm 1.0}$ & $\textcolor{blue}{22.6 \pm 0.7}$ & $\textcolor{blue}{12.0 \pm 0.4}$ & $\textcolor{blue}{6.2 \pm 0.2}$ \\

\hline
coil20 (H) & $46.8 \pm 1.4$ & $23.5 \pm 0.5$ & $11.8 \pm 0.5$ & $5.9 \pm 0.2$ \\
kylberg (H) & $\textcolor{blue}{46.9 \pm 2.2}$ & $\textcolor{blue}{23.9 \pm 0.3}$ & $\textcolor{blue}{12.0 \pm 0.04}$ & $\textcolor{blue}{6.0 \pm 0.04}$ \\

\hline
Dataset & \multicolumn{4}{c}{Compression Ratio} \\
(many classes) & 10\% & 20\% & 30\% & 40\% \\
\hline
{\sc{feret}} (C) & $9.9 \pm 0.1$ &	$\textcolor{blue}{5.0 \pm 0.1}$ & $\textcolor{blue}{3.3 \pm 0.006}$ & $\textcolor{blue}{2.5 \pm 0.01}$ \\
\hline
mpeg7 (H) & $9.9 \pm 0.2$ &	$4.9 \pm 0.06$ & $3.2 \pm 0.01$ & $2.4 \pm 0.04$ \\
\hline
\end{tabular}
}
\end{sc}
\end{small}
\end{table}

%% file: related.tex

Training set reduction has been considered in the context of {\knn} for vector data and the Euclidian distance with three primary methods: (a) \emph{training consistent sampling}, (b) \emph{prototype generation} and (c) \emph{prototype positioning} (for a survey see \cite{toussaint2002proximity}).
Training consistent sampling iteratively adds inputs from the training set to a reduced `reference set' until the reference set is perfectly classified by the training set.
This is precisely the technique of Condensed Nearest Neighbors (CNN) \cite{hart1968condensed}.
There have been a number of extensions of CNN, 
notably Reduced Nearest Neighbor (RNN) \cite{gates1972reduced} which searches for the smallest subset of the result of CNN that correctly classifies the training data.
Additionally, Fast CNN (FCNN) \cite{icml2005_Angiulli05} finds a set close to that of CNN but has training time linear (instead of cubic) in the size of the training set. Prototype generation creates new inputs to represent the training set, usually via clustering \cite{kohonen1990improved, salzberg1995best}. Prototype positioning learns a reduced training set by optimizing an appropriate objective. The method most similar to SCC and SHC is the recently proposed 
Stochastic Neighbor Compression \cite{kusner2014stochastic}, which uses a stochastic neighborhood to learn prototypes in Euclidean space (and thus is unsuitable for covariance and histogram features).
Finally, Bucilua et al.~\cite{bucilua2006model} may have been the first to study model compression for machine learning algorithms by compressing neural networks. 
To our knowledge, SCC and SHC are the first methods to explicitly consider training set reduction for SPD covariance and histogram descriptors.

Work towards speeding up test-time classification for {\knn} on covariance-valued data is somewhat limited. The JBLD divergence is proposed to speed up individual distance computations. Cherian et al.~\cite{cherian2011efficient} show that it is possible to adapt Bregman Ball Trees (BBTs), a generalization of the Euclidean ball tree to Bregman divergences, to the JBLD divergence. This is done using a clever iterative $K$-means method followed by a leaf node projection technique onto relevant Bregman balls. Both of these techniques are complementary to our dataset compression method.

There has been a large amount of work devoted toward improving the complexity of the Earth Mover's distance using approximations \cite{cuturi2013sinkhorn,grauman2004fast,shirdhonkar2008approximate,levina2001earth,ling2007efficient,pele2009fast}. For instance, \cite{pele2009fast} point out that if an upper bound can be placed on the transport $\Tb_{ij}$ between any two bins $i$ and $j$, then the (thresholded) EMD can be solved much more efficiently. Ling and Okada~\cite{ling2007efficient} show that if the ground distance is the $\ell_1$ distance between bins, the EMD can be reformulated exactly as a tree-based optimization problem with $d$ unknown variables (instead of $d^2$) and only $d$ constraints (instead of $2d$). We use the Sinkhorn approximation~\cite{cuturi2013sinkhorn}, which has the added advantage of an unconstrained dual formulation. 

%% file: conclusion.tex
In many classification settings the sheer amount of distance computations has previously prohibited the use of {\knn} for covariance and histogram features. We have shown that these data sets can be compressed to a small fraction of their original sizes while often only slightly increasing the test error. 
This drastically speeds up nearest neighbor search and has the potential to unlock new applications for covariance and histogram features on large datasets.

